\documentclass{article}
\usepackage{amsmath,epsfig}
\usepackage[preprint]{spconfa4}
\usepackage[misc]{ifsym}
\usepackage{multirow}
\usepackage{multicol}
\usepackage[american]{babel}
\usepackage{microtype}
\graphicspath{{Figure/}}
\usepackage{booktabs,bigstrut}
\usepackage{amsfonts}
\usepackage{caption}
\usepackage{subfigure}

\copyrightnotice{978-1-6654-3864-3/21/\$31.00 ©2021 IEEE}

\let\OLDthebibliography\thebibliography
\renewcommand\thebibliography[1]{
  \OLDthebibliography{#1}
  \setlength{\parskip}{0pt}
  \setlength{\itemsep}{0pt plus 0.3ex}
}

\begin{document}\sloppy

% Example definitions.
% --------------------
\def\x{{\mathbf x}}
\def\L{{\cal L}}

% Title.
% ------
\title{FGF-GAN: A Lightweight Generative Adversarial Network For Pansharpening Via Fast Guided Filter}
%
% Address.
% ---------------
\name{Zixiang Zhao, Jiangshe Zhang\textsuperscript{\Letter},\thanks{\textsuperscript{\Letter} Corresponding author. Email: jszhang@mail.xjtu.edu.cn}\thanks{The research is supported by the National Key Research and Development Program of China under grant 2018AAA0102201, the National Natural Science Foundation of China under grant 61976174 and 61877049.} 
	Shuang Xu, Kai Sun, Lu Huang, Junmin Liu, Chunxia Zhang}
\address{School of Mathematics and Statistics, Xi'an Jiaotong University, Xi’an 710049, China}

\maketitle

\begin{abstract}
	Pansharpening is a widely used image enhancement technique for remote sensing. Its principle is to fuse the input high-resolution single-channel panchromatic (PAN) image and low-resolution multi-spectral image and to obtain a high-resolution multi-spectral (HRMS) image. 
	The existing deep learning pansharpening method has two shortcomings. First, features of two input images need to be concatenated along the channel dimension to reconstruct the HRMS image, which makes the importance of PAN images not prominent, and also leads to high computational cost. Second, the implicit information of features is difficult to extract through the manually designed loss function.
	To this end, we propose a generative adversarial network via the fast guided filter (FGF) for pansharpening. In generator, traditional channel concatenation is replaced by FGF to better retain the spatial information while reducing the number of parameters. Meanwhile, the fusion objects can be highlighted by the spatial attention module. 
	In addition, the latent information of features can be preserved effectively through adversarial training.
	Numerous experiments illustrate that our network generates high-quality HRMS images that can surpass existing methods, and with fewer parameters.
\end{abstract}
\section{Introduction}\label{sec:intro}
Pansharpening (also known as remote sensing image fusion) is a hot issue in environmental monitoring and multi-modal image fusion.
The purpose of this task is to fuse the spatial/spectral information from source images including high-resolution single-channel panchromatic~(PAN) images and low-resolution multi-spectral (LRMS) images.
In the end, high-resolution multi-spectral (HRMS) images with the same size as PAN images and the same channel as LRMS images are obtained.

The traditional pansharpening method can be divided into three categories: component substitute~\cite{GILLESPIE1986209,laben2000process,haydn1982application}, multi-resolution analysis~\cite{DBLP:journals/lgrs/KhanCCM08,liu2000smoothing} and optimization-based methods~\cite{DBLP:journals/tgrs/GarzelliNC08}, where HRMS images are obtained by decomposing spectral/spatial information, injecting detail information or solving optimization models, respectively.
In the era of deep learning~(DL), DNN is often used as an extractor of spatial/spectral information and a fusion operator among multi-modal features. As a pioneering work, PNN~\cite{DBLP:journals/remotesensing/MasiCVS16} borrows a super-resolution convolutional neural network (CNN) to solve this task, but the network is relatively shallow and it is difficult to extract features effectively. DRPNN~\cite{DBLP:journals/lgrs/WeiYSZ17}, MSDCNN~\cite{DBLP:journals/staeors/YuanWMSZ18} and RSIFNN~\cite{DBLP:journals/staeors/ShaoC18} exploit deep residual learning, multi-scale and multi-depth CNN, and two-branch CNN to enlarge the layer number of deep network architecture and improve the ability of feature extraction. MIPSM~\cite{DBLP:journals/staeors/LiuWZLZZP20} combines shallow-deep CNN and spectral discrimination-based detail injection module to acquire HRMS images retaining more spectral information.

For DL-based methods, there are two shortcomings worth mentioning. 
First, before reconstructing the HRMS image, the multi-channel PAN and LRMS features often need to be concatenated in the channel dimension to pass through the reconstructor. This not only makes the spatial information in the PAN image indistinctive, but also causes the high training cost due to a huge number of parameters.
Second, manually designed reconstruction loss, such as $\ell_2$-loss, is difficult to accurately extract implicit feature information.

Therefore, we propose a lightweight generative adversarial network for pansharpening based on the fast guided filter~(FGF-GAN). In this paper, we formulate the pansharpening task as an adversarial training game, where the generator reconstructs a HRMS image containing spectral/spatial information from the LRMS/PAN image, and the discriminator distinguishes whether the generated image is a real sample.
Our contribution can be divided into three-fold:

(1) For the generator, to the best of our knowledge, this is the first time that fast guided filter~(FGF) is exploited to fuse the extracted feature maps in the pansharpening task. Compared with channel concatenation operation, the FGF method can better highlight the role of PAN image by setting the PAN image as the guidance image, while significantly reducing the number of parameters. In addition, cooperating with the spatial attention mechanism, object features useful for fusion are further concerned and the spectral/detailed texture information of the source image can be well preserved.

(2) Inspired by LSGAN~\cite{DBLP:conf/iccv/MaoLXLWS17}, adversarial training makes the generated images contain more latent details which are not easy to be learned by supervision of reconstruction loss.

(3) Extensive qualitative and quantitative experiments illustrate that our method can generate high-quality pansharpening images with fewer parameters.
\section{RELATED WORK}\label{sec:rw}
The guided filter~\cite{DBLP:journals/pami/He0T13}, as an anisotropic diffusion-based edge-aware image filter, is one of the fastest edge-preserving filters and is widely used in image enhancement/dehazing, HDR compression, saliency detection, etc.
For relatively ``flat'' patches in the guidance image where the variance is small, it can be regarded as a mean filter for input images in the corresponding area. For patches existing edges, it is equivalent to an edge-preserving filter. 
Subsequently, FGF~\cite{DBLP:journals/corr/He015} is proposed with an extra input, i.e., downsampling guidance image. So the smoothed maps are calculated on the low-resolution feature maps, which further improved the speed of the filter algorithm without losing accuracy.

Recently, Goodfellow et al.~\cite{goodfellow2014generative} propose a generative adversarial network (GAN) to complete the latent distribution learning of target data without any approximation through the adversarial training between generator and discriminator. 
LSGAN~\cite{DBLP:conf/iccv/MaoLXLWS17}, as a variant of GAN, replaces the cross-entropy loss of discriminator with least square loss function to solve the two shortcomings of traditional GAN, i.e., low quality for generated image and unstable training process. The idea of adversarial training is also applied in some information fusion tasks~\cite{DBLP:journals/inffus/MaYLLJ19,DBLP:journals/inffus/ZhangLSXM21}.
\section{Method}
In this section, we will introduce the network architecture and detailed information for our proposed FGF-GAN model.
\subsection{Motivation and Overview of FGF-GAN}
In order to better preserve the spectral and resolution information of the PAN/MS image, and meanwhile reduce the expensive calculation cost in the pansharpening task, we embed FGF~\cite{DBLP:journals/corr/He015} in the information fusion framework.
Different from the traditional applications for FGF which implement edge-aware in the image domain, we combine deep learning and FGF together, i.e., FGF is employed to transfer the information from PAN images into multi-spectral (MS) images in the feature domain.
By using the low-resolution feature maps instead of the full-resolution maps, FGF can significantly reduce the number of parameters while retaining the information fusion performance of our model.
In addition, some features that are difficult to be learned explicitly will be complemented by adversarial training.

\begin{figure*}[t]
	\centering
	\includegraphics[width=0.78\linewidth]{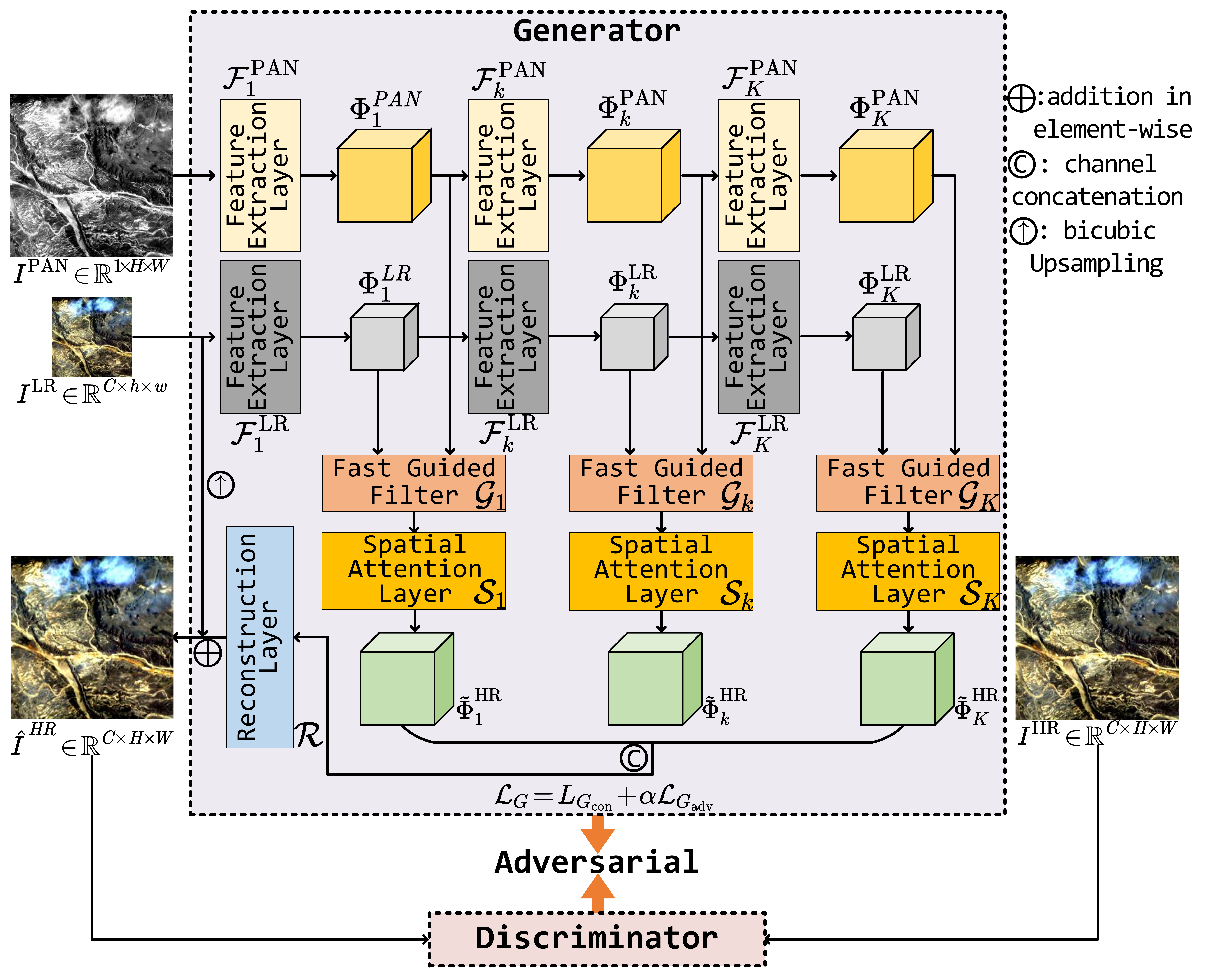}
	\caption{Neural network framework of FGF-GAN.}
	\label{fig:DFGF-GAN}
\end{figure*}

In short, our method implements PAN/LRMS image feature extraction, information fusion and HRMS image reconstruction through the generator, and then the supplement of extra information is accomplished through the discriminator.
For convenience, we define $I^\mathrm{PAN}\!\in\!\mathbb{R}^{1\!\times\!H\!\times\!W}, I^\mathrm{LR}\!\in\!\mathbb{R}^\mathit{C\!\times\!h\!\times\!w},I^\mathrm{HR}\!\in\!\mathbb{R}^\mathit{C\!\times\!H\!\times\!W},\hat{I}^\mathit{HR}\!\in\!\mathbb{R}^\mathit{C\!\times\!H\!\times\!W}$ as the input PAN image, the input LRMS image, the ground truth HRMS image and the output HRMS image of our model, respectively.
The framework of FGF-GAN is illustrated in Fig.~\ref{fig:DFGF-GAN} and the architecture in each module can be found in Tab.~\ref{tab:net_arc}.
\begin{table}[t]
	\centering
	\caption{The architecture of the FGF-GAN. Feature maps in Landsat8 is employed as samples to show the activation size.}
	\label{tab:net_arc}
	\begin{tabular}{ccc}
		\toprule
		Name                      &                 Layer$^{\rm *}$                 &        Activation size$^{\rm \dagger}$         \\ \midrule
		\multicolumn{3}{c}{\textbf{Generator Architecture}}                                                \\ \midrule
		$\mathcal{F}_1^\mathrm{PAN}$          &      $32\!\times\!1\!\times\!3$ conv, ReLU      &          $64\!\times\!64\!\times\!32$          \\
		$\mathcal{F}_1^\mathrm{LR}$          &     $32\!\times\! C\!\times\!3$ conv, ReLU      &          $32\!\times\!32\!\times\!32$          \\
		\multirow{2}{*}{$\mathcal{F}_k^\mathrm{PAN}$$^{\rm \ddagger}$} &        $32\!\times\!32\!\times\!3$ conv,        & \multirow{2}{*}{$64\!\times\!64\!\times\!32$}  \\
		&     $32\!\times\!32\!\times\!3$ conv, ReLU      &                                                \\
		\multirow{2}{*}{$\mathcal{F}_k^\mathrm{LR}$}  &        $32\!\times\!32\!\times\!3$ conv,        & \multirow{2}{*}{$32\!\times\!32\!\times\!32$}  \\
		&     $32\!\times\!32\!\times\!3$ conv, ReLU      &                                                \\
		$\mathcal{G}_k$                &                        -                        &          $64\!\times\!64\!\times\!32$          \\
		$\mathcal{S}_k$                &   SAM in CBAM~\cite{DBLP:conf/eccv/WooPLK18}    &          $64\!\times\!64\!\times\!32$          \\
		\multirow{2}{*}{$ \mathcal{R} $}        & $32\!\times\!(32\!\times\! K)\!\times\!3$ conv, & \multirow{2}{*}{ $64\!\times\!64\!\times\! C$} \\
		&      ReLU, $C\!\times\!32\!\times\!3$ conv      &                                                \\ \midrule
		\multicolumn{3}{c}{\textbf{Discriminator Architecture}}                                              \\ \midrule
		\multicolumn{2}{c}{$32\!\times\!(2\!\times\! C+1)\!\times\!3$ conv, BN, LReLU}          &           $8\!\times\!8\!\times\!32$           \\
		\multicolumn{2}{c}{$64\!\times\!32\!\times\!3$ conv, BN, LReLU }                 &           $8\!\times\!8\!\times\!64$           \\
		\multicolumn{2}{c}{$128\!\times\!64\!\times\!3$ conv, BN, LReLU }                &          $8\!\times\!8\!\times\!128$           \\
		\multicolumn{2}{c}{$256\!\times\!128\!\times\!3$ conv, BN, LReLU }                &          $8\!\times\!8\!\times\!256$           \\
		\multicolumn{2}{c}{$1\!\times\!256\!\times\!3$ conv, BN, Sigmoid }                &           $8\!\times\!8\!\times\!1$            \\ \bottomrule
	\end{tabular}
	\leftline{\footnotesize{$^*$ Convolution kernels: output channels$\times$input channels$\times$kernel size.}}
	\leftline{\footnotesize{$^\dagger$ The representation paradigm: width $\times$ height $\times$ channels.}}
	\leftline{\footnotesize{$^{\rm \ddagger}$ $k=2,3,\cdots,K$ in $\mathcal{F}_k^\mathrm{PAN}$ and $\mathcal{F}_k^\mathrm{LR}$ here.}}
\end{table}
\subsection{Generator Details}
There are four main components in the generator of FGF-GAN, i.e., feature extraction layer, FGF layer, spatial attention layer and image reconstruction layer. The goal of each module is to capture features useful for information fusion from the source images, fuse and retain the extracted features, highlight the salient objects and obtain the HRMS images, respectively.

In detail, the paired $\{I^\mathrm{PAN}_n,I^\mathrm{LR}_n\}_{n=1}^N$ are input into feature extractor $\{\mathcal{F}_k^\mathrm{PAN},\mathcal{F}_k^\mathrm{LR}\}_{k=1}^K$, where $\{\mathcal{F}_k^\mathrm{PAN},\mathcal{F}_k^\mathrm{LR}\}$ denote the $k$th feature extraction layer for PAN/LRMS images in the generator. Note that there is no shared parameter between $\mathcal{F}_k^\mathrm{PAN}$ and $\mathcal{F}_k^\mathrm{LR}$, and the number of feature extraction layers $K$ is determined in the validation set. After obtaining the corresponding feature maps $\{\Phi_k^\mathrm{PAN},\Phi_k^\mathrm{LR}\}_{k=1}^K$, the information fusion is accomplished by
\begin{equation}\label{equ:FGF}
\Phi_k^\mathrm{HR}=\mathcal{G}_k \left (  {\Phi_k^\mathrm{PAN}\!\downarrow} ,\Phi_k^\mathrm{LR}, \Phi_k^\mathrm{PAN} \right ), 
\end{equation}
where $\mathcal{G}_k(\cdot,\cdot,\cdot)$ represented the FGF operator~\cite{DBLP:journals/corr/He015} corresponding to the $k$th group of feature maps and $\downarrow$ is the bicubic downsampling operator. Then we exploit the spatial attention layer $\mathcal{S}_k$ to highlight the salient objects for fusion by:
\begin{equation}\label{equ:SAL}
\tilde{\Phi}_k^\mathrm{HR}=\mathcal{S}_k \left ( \Phi_k^\mathrm{HR}\right ),
\end{equation}  
where $\mathcal{S}_k$ is the spatial attention module (SAM) in CBAM~\cite{DBLP:conf/eccv/WooPLK18}.	
At last, the attention feature maps $\{\tilde{\Phi}_k^\mathrm{HR}\}_{k=1}^K$ are concatenated along the channel dimension and are input into the reconstruction layer $\mathcal{R}(\cdot)$. 
At last, it sets a skip connection between the upsampled LRMS image and the output of the reconstruction layer to output the final HRMS image, that is,
\begin{equation}\label{equ:IRL}
\hat{I}^\mathrm{HR}=\mathcal{R}(\tilde{\Phi}^\mathrm{HR})\oplus (I^\mathrm{LR}\uparrow),
\end{equation}
where $\uparrow$ is the bicubic upsampling operator and $\oplus$ is the element-wise addition.
\subsection{Discriminator Details}
After obtaining the $\hat{I}^\mathrm{HR}$, in order to further enhance the texture information and implicit detail information of fusion images, we establish an adversarial game between the generator and the discriminator. We input $\{{I}^\mathrm{HR},\hat{I}^\mathrm{HR}\}$ into the discriminator, and define ${I}^\mathrm{HR}$ as real data while $\hat{I}^\mathrm{HR}$ as fake data.
Through the adversarial training, the quality of fusion images is effectively improved, and the rationality of adversarial training is proved in the ablation experiment in Sec.~\ref{sec:Ablation}.
\subsection{Loss Function}
The loss function of our FGF-GAN consists of two components, the loss of generator $\mathcal{L}_{G}$ and the loss of discriminator $\mathcal{L}_{D}$. In the following, we will introduce the two components respectively.\\
\textbf{Loss function of generator.}
There are two terms in $\mathcal{L}_{G}$, i.e., the reconstruct loss $\mathcal{L}_{G_{\mathrm{con}}}$ and the adversarial loss $\mathcal{L}_{G_{\mathrm{adv}}}$. For $\mathcal{L}_{G_{\mathrm{con}}}$, the $\ell_1$-loss is selected to make our model robust to the outlier in regression. For $\mathcal{L}_{G_{\mathrm{adv}}}$, inspired by LSGAN~\cite{DBLP:conf/iccv/MaoLXLWS17}, the $\ell_2$-loss between the predicted probability and the real data label is used to make the discriminator $D_{\theta_D}$ believe for the generated fake data. So $\mathcal{L}_{G}$ is formulated as:
\begin{equation}\label{eq:L_G}
\begin{split}
\mathcal{L}_{G}\!=&\!L_{G_{\mathrm{con}}}\!+\!\alpha \mathcal{L}_{G_{\mathrm{adv}}} \\
\!=&\sum_{n=1}^{N}\left \| {I}_n^\mathrm{HR},\hat{I}_n^\mathrm{HR} \right \|_1 +\!\frac{\alpha}{N} \sum_{n=1}^{N}\left(D_{\theta_D}(\hat{I}_n^\mathit{HR})\!-\!a\right)^{2},
\end{split}
\end{equation}
where $a$ is the real data label and $\alpha$ is the turning parameter.\\
\textbf{Loss function of discriminator.} The discriminator is established to distinguish the ground truth and the generated HRMS image, and $\mathcal{L}_{D}$ is expressed as: 
\begin{equation}\label{eq:L_D}
\mathcal{L}_{D_{\mathrm{adv}}}\!=\!\frac{1}{N}\!\sum_{n=1}^{N}\left[D_{\theta_D}\!\left(\hat{I}_n^\mathit{HR}\right)\!-\!b\right]^{2}\!+\!\left[D_{\theta_D}\!\left({I}_n^\mathit{HR}\right)\!-\!c\right]^{2},
\end{equation}
where $b$ and $c$ are labels of fake data and real data, respectively.
\section{Experiments}
In this section, we will use a variety of experiments to verify the effectiveness of our model and the rationality of the module settings. All experiments are conducted with Pytorch on a computer with Intel Core i9-10900K CPU@3.70GHz and NVIDIA GeForce RTX2080Ti GPU.
\begin{table}[t]
	\centering
	\caption{Dataset employed in this paper.}
	\label{tab:dataset}
	\begin{tabular}{cccc}
		\toprule
		Dataset& Division$^{\rm *}$    & Bands & SUS ratio$^{\rm \dagger}$ \\\midrule
		Landsat8  & 350/50/100  & 10    & 2         \\
		QuickBird & 474/103/100 & 4     & 4         \\
		GaoFen2   & 350/50/100  & 4     & 4         \\
		\bottomrule
	\end{tabular}
	\leftline{\footnotesize{$^*$ The paradigm of dataset division: number of training/validation/test sets.}}
	\leftline{\footnotesize{$^\dagger$ SUS ratio represents the spatial up-scaling ratio.}}
\end{table}
\begin{figure*}[t]
	\centering
	\subfigure[PAN]{\includegraphics[width=0.14\linewidth]{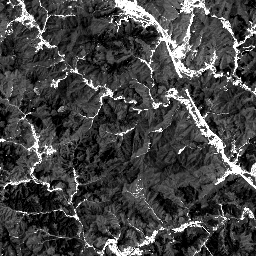}}\hspace{-0.2em}
	\subfigure[LRMS]{\includegraphics[width=0.14\linewidth]{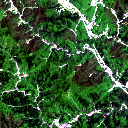}}\hspace{-0.2em}
	\subfigure[Ground Truth]{\includegraphics[width=0.14\linewidth]{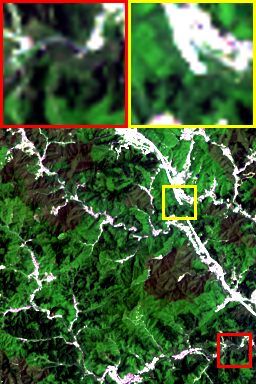}}\hspace{-0.2em}
	%		\subfigure[BDSD~\cite{DBLP:journals/tgrs/GarzelliNC08}]{\includegraphics[width=0.14\linewidth]{qualitative/Landsat8-97-BDSD}}\hspace{-0.2em}
	\subfigure[Brovey~\cite{GILLESPIE1986209}]{\includegraphics[width=0.14\linewidth]{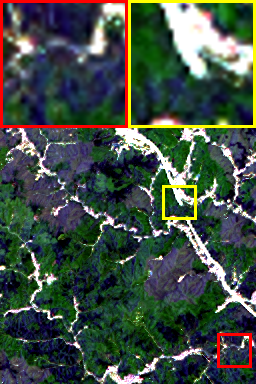}}\hspace{-0.2em}
	\subfigure[GS~\cite{laben2000process}]{\includegraphics[width=0.14\linewidth]{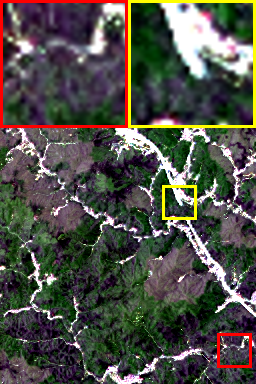}}\hspace{-0.2em}
	\subfigure[HPF~\cite{schowengerdt1980reconstruction}]{\includegraphics[width=0.14\linewidth]{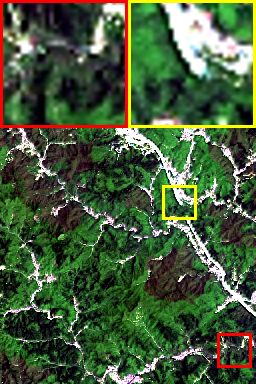}}\hspace{-0.2em}
	\subfigure[IHS~\cite{haydn1982application}]{\includegraphics[width=0.14\linewidth]{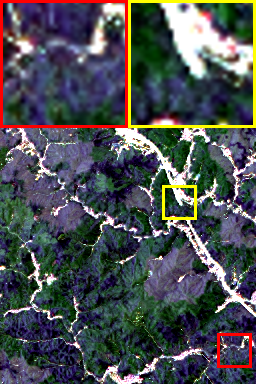}}\hspace{-0.2em}\\\vspace{-1em}
	\subfigure[Indusion~\cite{DBLP:journals/lgrs/KhanCCM08}]{\includegraphics[width=0.14\linewidth]{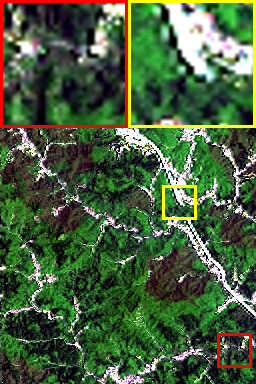}}\hspace{-0.2em}
	\subfigure[SFIM~\cite{liu2000smoothing}]{\includegraphics[width=0.14\linewidth]{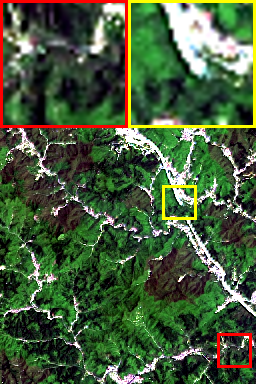}}\hspace{-0.2em}
	\subfigure[MIPSM~\cite{DBLP:journals/staeors/LiuWZLZZP20}]{\includegraphics[width=0.14\linewidth]{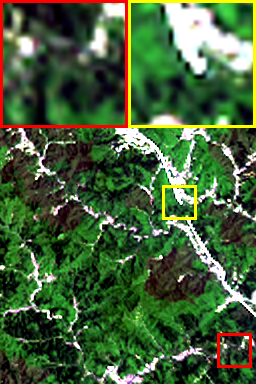}}\hspace{-0.2em}
	\subfigure[DRPNN~\cite{DBLP:journals/lgrs/WeiYSZ17}]{\includegraphics[width=0.14\linewidth]{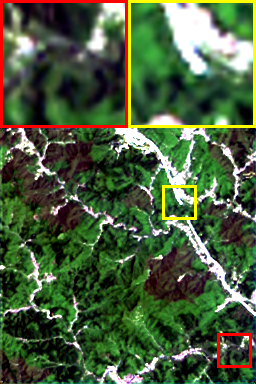}}\hspace{-0.2em}
	\subfigure[MSDCNN~\cite{DBLP:journals/staeors/YuanWMSZ18}]{\includegraphics[width=0.14\linewidth]{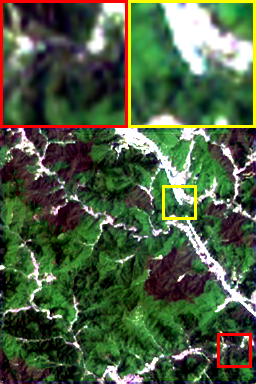}}\hspace{-0.2em}
	\subfigure[RSIFNN~\cite{DBLP:journals/staeors/ShaoC18}]{\includegraphics[width=0.14\linewidth]{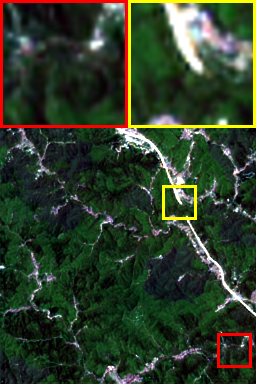}}\hspace{-0.2em}
	\subfigure[FGF-GAN (Ours)]{\includegraphics[width=0.14\linewidth]{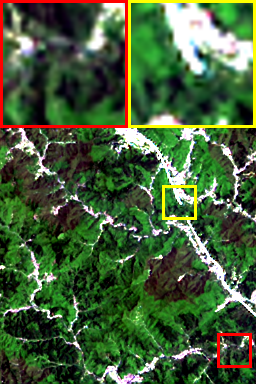}}\hspace{-0.2em}
	\caption{Qualitative results for SOTAs and ours.}
	\label{fig:DFGF-GAN1}
\end{figure*}
\begin{table*}[!]
	\centering
	\caption{Quantitative results. \textbf{Bold} indicates the best result and \underline{underlined} indicates the second best result.}
	\label{tab:test}
	\resizebox{0.99\linewidth}{!}{
	\begin{tabular}{c|cccc|cccc|cccc}
		\toprule
		&                \multicolumn{4}{c|}{\textbf{Dataset: Landsat8}}                 &                \multicolumn{4}{c|}{\textbf{Dataset: QuickBird}}                &                 \multicolumn{4}{c}{\textbf{Dataset: GaoFen2}}                  \\
		Methods  &        PSNR        &        CC         &        SAM        &       ERGAS       &        PSNR        &        CC         &        SAM        &       ERGAS       &        PSNR        &        CC         &        SAM        &       ERGAS       \\ \midrule
		BDSD   &       33.806       &       0.889       &       0.025       &       1.913       &       23.554       &       0.737       &       0.076       &       4.887       &       30.211       &       0.912       &       0.013       &       2.396       \\
		Brovey  &       32.403       &       0.703       &       0.021       &       1.981       &       25.274       &       0.823       &       0.064       &       4.209       &       31.590       &       0.952       &       0.011       &       2.209       \\
		GS    &       32.016       &       0.815       &       0.030       &       2.212       &       26.031       &       0.686       &       0.059       &       3.950       &       30.436       &       0.916       &       0.010       &       2.308       \\
		HPF    &       32.669       &       0.843       &       0.025       &       2.067       &       25.998       &       0.789       &       0.059       &       3.945       &       30.481       &       0.923       &       0.011       &       2.331       \\
		IHS    &       32.877       &       0.732       &       0.024       &       2.313       &       24.383       &       0.751       &       0.065       &       4.621       &       30.475       &       0.924       &       0.011       &       2.355       \\
		Indusion &       30.848       &       0.758       &       0.036       &       2.422       &       25.762       &       0.651       &       0.067       &       4.251       &       30.536       &       0.920       &       0.011       &       2.346       \\
		SFIM   &       32.721       &       0.843       &       0.025       &       2.078       &       24.035       &       0.676       &       0.074       &       4.828       &       30.402       &       0.910       &       0.013       &       2.369       \\
		MIPSM   &       35.489       & \underline{0.891} &       0.021       &       1.577       &       27.732       &       0.841       &       0.052       &       3.155       &       32.176       &       0.955       &       0.010       &       1.883       \\
		DRPNN   & \underline{37.364} &       0.889       &       0.017       &       1.330       & \underline{31.041} &  \textbf{0.910}   &  \textbf{0.038}   & \underline{2.225} & \underline{35.118} &       0.971       &       0.010       & \underline{1.278} \\
		MSDCNN  &       36.254       &       0.887       &       0.018       &       1.416       &       30.124       &       0.889       &       0.043       &       2.565       &       33.671       &  \textbf{0.979}   & \underline{0.009} &       1.472       \\
		RSIFNN  &       37.078       &       0.884       & \underline{0.017} & \underline{1.327} &       30.577       &       0.890       &       0.040       &       2.353       &       33.059       & {0.972} &       0.011       &       1.566       \\\midrule
		Ours   &  \textbf{38.179}   &  \textbf{0.902}   &  \textbf{0.016}   &  \textbf{1.271}   &  \textbf{31.458}   & \underline{0.906} & \underline{0.038} &  \textbf{2.185}   &  \textbf{35.193}   &       \underline{0.972}       &  \textbf{0.009}   &  \textbf{1.276}   \\ \bottomrule
	\end{tabular}}
\end{table*}
\newcommand{\widthzzx}{0.877}
\begin{figure*}[t]
	\centering
	\begin{minipage}[t]{0.33\textwidth}
		\centering
		\includegraphics[width=\widthzzx\linewidth]{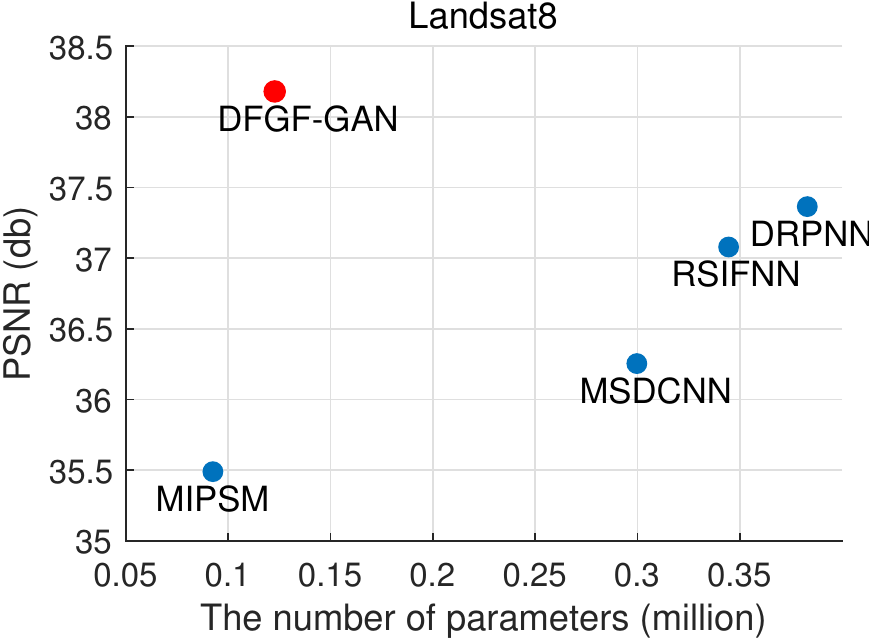}
		\caption*{(a) Landsat8}
		\par\vspace{0pt}
	\end{minipage}
	\begin{minipage}[t]{0.33\linewidth}
		\centering
		\includegraphics[width=\widthzzx\linewidth]{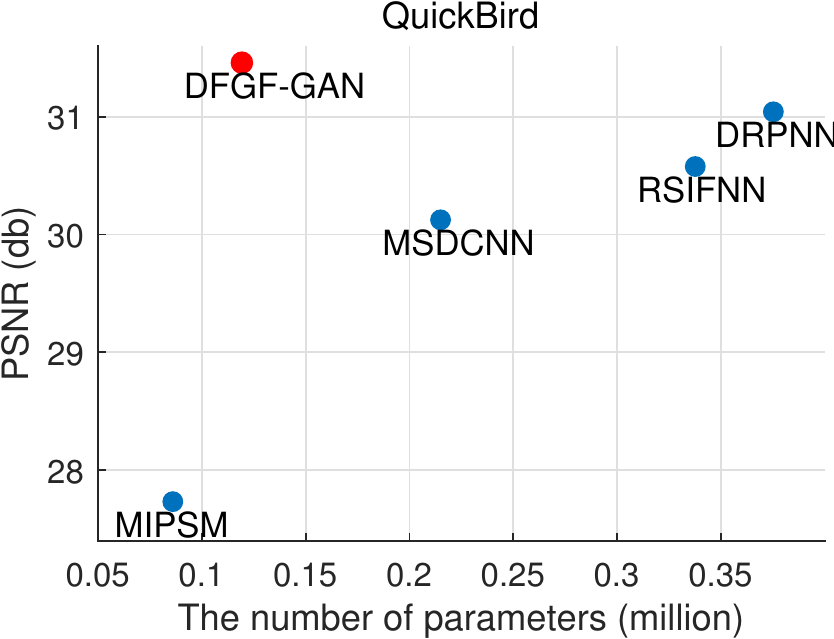}
		\caption*{(b) QuickBird}
		\par\vspace{0pt}
	\end{minipage}
	\begin{minipage}[t]{0.33\linewidth}
		\centering
		\includegraphics[width=\widthzzx\linewidth]{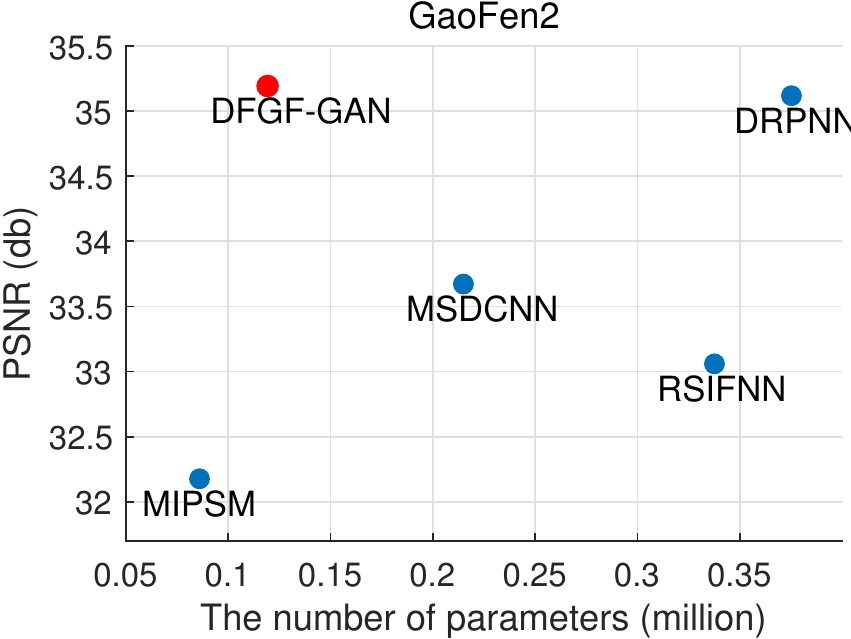}
		\caption*{(c) GaoFen2}
		\par\vspace{0pt}
	\end{minipage}
	\caption{Parameters comparison results in test sets of Landsat8, QuickBird and GaoFen2, respectively.}
	\label{fig:parameter}
\end{figure*}
\begin{table*}[t]
	\centering
	\caption{Results of the determination of feature extraction layer number $K$ (left) and ablation experiments (right) in the validation set and test set of Landsat8, respectively. \textbf{Bold} indicates the best result.}
	\label{tab:Ablation}
	\resizebox{0.9\linewidth}{!}{
	\begin{tabular}{c|cccccc|ccc}
		\toprule
		& \multicolumn{6}{c|}{\textbf{Determination of feature extraction layer number $K$}} & \multicolumn{3}{c}{\textbf{Ablation Experiments}} \\
		Metrics  &    2     &    3     &    4     &    5     &    6     &     7      &       Ours        & w/o GAN  &      w/o SAM       \\ \midrule
		%		 PSNR  & 37.75476 & 37.88311 & 38.17860 & 38.19587 & 38.20393 &  38.21397  & \textbf{38.11950} & 37.07674 &      37.19206      \\
		%		  CC  & 0.89562  & 0.90078  & 0.90231  & 0.90234  & 0.90231  &  0.90229   & \textbf{0.90850}  & 0.90151  &      0.88081       \\
		%		  SAM  & 0.01722  & 0.01686  & 0.01633  & 0.01628  & 0.01627  &  0.01622   & \textbf{0.01645}  & 0.01868  &      0.01860       \\
		%		 ERGAS & 1.32488  & 1.31114  & 1.27061  & 1.26957  & 1.26867  &  1.26863   & \textbf{1.25824}  & 1.29619  &      1.37268       \\
		PSNR    & 37.69566 & 37.82401 & 38.11950 & 38.13677 & 38.14483 &  38.15397  & \textbf{38.17860} & 37.07674 &      37.19206      \\
		CC     & 0.90181  & 0.90697  & 0.90850  & 0.90853  & 0.90850  &  0.90849   & \textbf{0.90231}  & 0.90151  &      0.88081       \\
		SAM    & 0.01735  & 0.01699  & 0.01645  & 0.01641  & 0.01639  &  0.01638   & \textbf{0.01633}  & 0.01868  &      0.01860       \\
		ERGAS   & 1.30251  & 1.27877  & 1.25824  & 1.25720  & 1.25630  &  1.25563   & \textbf{1.27061}  & 1.29619  &      1.37268       \\ \bottomrule
	\end{tabular}}
\end{table*}
\subsection{Experimental Settings}
\textbf{Datasets and evaluation metrics.} We choose three satellite datasets, i.e., Landsat8, QuickBird and GaoFen2, as our experimental datasets, whose details are displayed in Tab.~\ref{tab:dataset}. Wald protocol~\cite{Wald97fusionof} is employed to prepare for the training samples. Additionally, in the training phase, we crop the LRMS samples into 32$\times$32 patches and PAN samples into (32$\times$SUS)$\times$(32$\times$SUS) patches, where SUS denotes the spatial up-scaling ratio.
In the test phase, three spatial assessment metrics including peak signal-to-noise ratio (PSNR), correlation coefficient~(CC), relative dimensionless global error in synthesis~(ERGAS) and a spectral assessment metric spectral angle mapper~(SAM) are employed to measure the fusion effectiveness of our model. The larger PSNR, CC and the smaller ERGAS, SAM imply the higher quality of the fusion images.\\	
\textbf{Training details.}
In the training phase, FGF-GAN is optimized by Adam in 200 epochs. The batch size is set to 64 and the learning rate equals to $5\!\times\!10^{-4}$ with a 10 times decrease after 100 epochs. As for the data labels in Eqs.~(\ref{eq:L_G}) and (\ref{eq:L_D}), we set $a,c=U(0.9,1.1)$ and $b=U(0,0.2)$ as soft labels~\cite{DBLP:conf/iccv/MaoLXLWS17}, where $U(\cdot,\cdot)$ denotes the uniform distribution. $\alpha$ is set to 0.01 in Eq.~(\ref{eq:L_G}) to keep loss terms with the same magnitude.

\subsection{Comparison with SOTAs}
In this section, we compare our FGF-GAN with some SOTA methods in pansharpening, including 
BDSD~\cite{DBLP:journals/tgrs/GarzelliNC08},
Brovey~\cite{GILLESPIE1986209},
GS~\cite{laben2000process},
HPF~\cite{schowengerdt1980reconstruction},
IHS~\cite{haydn1982application},
Indusion~\cite{DBLP:journals/lgrs/KhanCCM08},
SFIM~\cite{liu2000smoothing},
MIPSM~\cite{DBLP:journals/staeors/LiuWZLZZP20},
DRPNN~\cite{DBLP:journals/lgrs/WeiYSZ17},
MSDCNN~\cite{DBLP:journals/staeors/YuanWMSZ18} and 
RSIFNN~\cite{DBLP:journals/staeors/ShaoC18}.\\
\textbf{Qualitative comparison.} Visual inspection results of RGB bands are displayed in Fig.~\ref{fig:DFGF-GAN1}. Compared with other methods, FGF-GAN can better preserve the detailed texture information of the PAN images and the spectral information of the MS images, and the generated images are the closest to the ground truth. The amplified area can also show that our method successfully avoids spatial and spectral distortion, and can generate high-quality fusion results with lower noise.\\
\textbf{Quantitative comparison.} The values of metrics in three test datasets are exhibited in Tab.~\ref{tab:test}. It is obvious that our model achieves almost all the best results with regard to all metrics while the other methods can only perform well in a part of metrics. It proves that our method can generate excellent fusion images and preserve more texture details and spectral information from source images.\\
\textbf{Parameters comparison.} In addition, we compare the parameters contained in the DL-based models and exhibit them in Fig.~\ref{fig:parameter}. The result shows that our method can generate satisfactory pansharpening images with fewer parameters and demonstrates the superiority of our lightweight networks.

\subsection{Ablation Experiments}\label{sec:Ablation}
\textbf{Layer number determination.} The number of feature extraction layers $K$ of $\mathcal{F}_k^\mathrm{PAN}$ and $\mathcal{F}_k^\mathrm{LR}$ is a very important parameter in our model. We train models with different $K$ and show fusion results of validation set in Tab.~\ref{tab:Ablation}. When $K>4$, the fusion results are not significantly improved. Balancing model accuracy and computational cost, we finally determine $K=4$.\\
\textbf{Module rationality analysis.} We show the rationality of adversarial training and spatial attention layer by ablation experiments. In experiment \textit{w/o~GAN}, the adversarial training is eliminated, i.e., only the generator is trained to complete pansharpening. In the experiment \textit{w/o~SAM}, we deleted the spatial attention layer in the generator and display the results of Landsat8 test set in Tab.~\ref{tab:Ablation}. 
Compared with FGF-GAN, the fusion effects of the two ablation experiment groups are reduced, which proves the rationality of our model.

\section{Conclusion}
In this paper, we propose a lightweight pansharpening network based on generative adversarial training. In the generator, the fast guided filter is used to fuse the extracted feature maps. Compared with the existing methods that use channel concatenation before reconstruction, the fast guided filter can better emphasize the role of PAN images in pansharpening while reducing the number of parameters. 
In addition, the spatial attention mechanism highlights the targets that need to be fused, and the adversarial training makes the generated images contain more latent information from source images. Qualitative and quantitative results show that our method can generate satisfactory pansharpening images with fewer model parameters.

\bibliographystyle{IEEEbib}
\bibliography{icme}

\begin{thebibliography}{10}

\bibitem{GILLESPIE1986209}
A.~B.~Kahle A.~R.~Gillespie and R.~E. Walker,
\newblock ``Color enhancement of highly correlated images. i. decorrelation and
  hsi contrast stretches,''
\newblock {\em Remote Sens. of Environ.}, vol. 20, no. 3, pp. 209 -- 235, 1986.

\bibitem{laben2000process}
C.~A. Laben, V.~Bernard, and W.~Brower,
\newblock ``Process for enhancing the spatial resolution of multispectral
  imagery using pan-sharpening,'' Jan. 2000,
\newblock US Patent 6,011,875.

\bibitem{haydn1982application}
R.~Haydn, G.~W. Dalke, J.~Henkel, and J.~E. Bare,
\newblock ``Application of the ihs color transform to the processing of
  multisensor data and image enhancement,''
\newblock in {\em Int. Symp. on Remote Sens. of Environ.}, 1982.

\bibitem{DBLP:journals/lgrs/KhanCCM08}
M.~M. Khan, J.~Chanussot, L.~Condat, and A.~Montanvert,
\newblock ``Indusion: Fusion of multispectral and panchromatic images using the
  induction scaling technique,''
\newblock {\em {IEEE} GRSL}, vol. 5, no. 1, pp. 98--102, 2008.

\bibitem{liu2000smoothing}
J.~G. Liu,
\newblock ``Smoothing filter-based intensity modulation: A spectral preserve
  image fusion technique for improving spatial details,''
\newblock {\em Int. J. Remote Sens.}, vol. 21, no. 18, pp. 3461--3472, 2000.

\bibitem{DBLP:journals/tgrs/GarzelliNC08}
A.~Garzelli, F.~Nencini, and L.~Capobianco,
\newblock ``Optimal {MMSE} pan sharpening of very high resolution multispectral
  images,''
\newblock {\em {IEEE} TGRS}, vol. 46, no. 1, pp. 228--236, 2008.

\bibitem{DBLP:journals/remotesensing/MasiCVS16}
G.~Masi, D.~Cozzolino, L.~Verdoliva, and G.~Scarpa,
\newblock ``Pansharpening by convolutional neural networks,''
\newblock {\em Remote. Sens.}, vol. 8, no. 7, pp. 594, 2016.

\bibitem{DBLP:journals/lgrs/WeiYSZ17}
Y.~Wei, Q.~Yuan, H.~Shen, and L.~Zhang,
\newblock ``Boosting the accuracy of multispectral image pansharpening by
  learning a deep residual network,''
\newblock {\em {IEEE} GRSL}, vol. 14, no. 10, pp. 1795--1799, 2017.

\bibitem{DBLP:journals/staeors/YuanWMSZ18}
Q.~Yuan, Y.~Wei, X.~Meng, H.~Shen, and L.~Zhang,
\newblock ``A multiscale and multidepth convolutional neural network for remote
  sensing imagery pan-sharpening,''
\newblock {\em {IEEE} JSTARS}, vol. 11, no. 3, pp. 978--989, 2018.

\bibitem{DBLP:journals/staeors/ShaoC18}
Z.~Shao and J.~Cai,
\newblock ``Remote sensing image fusion with deep convolutional neural
  network,''
\newblock {\em {IEEE} JSTARS}, vol. 11, no. 5, pp. 1656--1669, 2018.

\bibitem{DBLP:journals/staeors/LiuWZLZZP20}
L.~Liu, J.~Wang, E.~Zhang, B.~Li, X.~Zhu, Y.~Zhang, and J.~Peng,
\newblock ``Shallow-deep convolutional network and
  spectral-discrimination-based detail injection for multispectral imagery
  pan-sharpening,''
\newblock {\em {IEEE} JSTARS}, vol. 13, pp. 1772--1783, 2020.

\bibitem{DBLP:conf/iccv/MaoLXLWS17}
X.~Mao, Q.~Li, H.~Xie, R.~Y.~K. Lau, Z.~Wang, and S.~P. Smolley,
\newblock ``Least squares generative adversarial networks,''
\newblock in {\em {ICCV}}, 2017, pp. 2813--2821.

\bibitem{DBLP:journals/pami/He0T13}
K.~He, J.~Sun, and X.~Tang,
\newblock ``Guided image filtering,''
\newblock {\em {IEEE} TPAMI}, vol. 35, no. 6, pp. 1397--1409, 2013.

\bibitem{DBLP:journals/corr/He015}
K.~He and J.~Sun,
\newblock ``Fast guided filter,''
\newblock {\em CoRR}, vol. abs/1505.00996, 2015.

\bibitem{goodfellow2014generative}
I.~Goodfellow, J.~Pouget-Abadie, M.~Mirza, B.~Xu, D.~Warde-Farley, S.~Ozair,
  A.~Courville, and Y.~Bengio,
\newblock ``Generative adversarial nets,''
\newblock in {\em NeurIPS}, 2014, pp. 2672--2680.

\bibitem{DBLP:journals/inffus/MaYLLJ19}
J.~Ma, W.~Yu, P.~Liang, C.~Li, and J.~Jiang,
\newblock ``Fusiongan: {A} generative adversarial network for infrared and
  visible image fusion,''
\newblock {\em Inf. Fusion}, vol. 48, pp. 11--26, 2019.

\bibitem{DBLP:journals/inffus/ZhangLSXM21}
H.~Zhang, Z.~Le, Z.~Shao, H.~Xu, and J.~Ma,
\newblock ``{MFF-GAN:} an unsupervised generative adversarial network with
  adaptive and gradient joint constraints for multi-focus image fusion,''
\newblock {\em Inf. Fusion}, vol. 66, pp. 40--53, 2021.

\bibitem{DBLP:conf/eccv/WooPLK18}
S.~Woo, J.~Park, J.~Lee, and I.~S. Kweon,
\newblock ``{CBAM:} convolutional block attention module,''
\newblock in {\em {ECCV}}, 2018, pp. 3--19.

\bibitem{schowengerdt1980reconstruction}
R.~A. Schowengerdt,
\newblock ``Reconstruction of multispatial, multispectral image data using
  spatial frequency content,''
\newblock {\em Photogramm. Eng. and Remote Sens.}, vol. 46, no. 10, pp.
  1325--1334, 1980.

\bibitem{Wald97fusionof}
L.~Wald, T.~Ranchin, and M.~Mangolini,
\newblock ``Fusion of satellite images of different spatial resolution:
  Assessing the quality of resulting images,''
\newblock {\em Photogramm. Eng. Remote Sens.}, pp. 691--699, 1997.

\end{thebibliography}

\end{document}